\documentclass[runningheads]{llncs}

\usepackage[T1]{fontenc}
\usepackage{graphicx}
\usepackage{booktabs}           
\usepackage{xcolor}             
\usepackage{listings}           
\usepackage{amsmath,amssymb}    
\usepackage{hyperref}           
\usepackage{cleveref}           
\usepackage{enumitem}
 
\hypersetup{
    colorlinks=true,
    linkcolor=blue,
    citecolor=blue,
    urlcolor=blue
}

\crefname{section}{Section}{Sections}
\crefname{figure}{Figure}{Figures}
\crefname{table}{Table}{Tables}
\crefname{equation}{Equation}{Equations}

\begin{document}

\title{From Research Question to Scientific Workflow:\\Leveraging Agentic AI for Science Automation}

\author{Bartosz Balis\inst{1} \and
Micha{\l} Orzechowski\inst{1,2} \and
Piotr Kica\inst{1,2} \and\\
Micha{\l} Dygas\inst{1} \and
Micha{\l} Kuszewski\inst{1} 
}

\authorrunning{B. Balis et al.}

\institute{AGH University of Krakow, Krakow, Poland\\
\email{\{balis,morzech,kica\}@agh.edu.pl}
\and Sano Centre for Computational Medicine, Krakow, Poland}

\maketitle

\begin{abstract}
Scientific workflow systems automate execution -- scheduling, fault tolerance, resource management -- but not the semantic translation that precedes it. Scientists still manually convert research questions into workflow specifications, a task requiring both domain knowledge and infrastructure expertise. We propose an agentic architecture that closes this gap through three layers: an LLM interprets natural language into structured intents (semantic layer); validated generators produce reproducible workflow DAGs (deterministic layer); and domain experts author ``Skills'': markdown documents encoding vocabulary mappings, parameter constraints, and optimization strategies (knowledge layer). This decomposition confines LLM non-determinism to intent extraction: identical intents always yield identical workflows. We implement and evaluate the architecture on the 1000 Genomes population genetics workflow and Hyperflow WMS running on Kubernetes. In an ablation study on 150 queries, Skills raise full-match intent accuracy from 44\% to 83\%; skill-driven deferred workflow generation reduces data transfer by 92\%; and the end-to-end pipeline completes queries on Kubernetes with LLM overhead below 15 seconds and cost under \$0.001 per query.

\end{abstract}


\section{Introduction}
\label{sec:introduction}

Scientific workflow management systems such as Pegasus~\cite{deelman2015pegasus}, Nextflow~\cite{ditommaso2017nextflow}, Snakemake~\cite{koster2012snakemake}, and Galaxy~\cite{afgan2018galaxy} automate the execution of computational ex\-pe\-ri\-ments \cite{suter2025terminology}, handling task scheduling, fault tolerance, data staging, and resource management across distributed infrastructures. Given a workflow specification -- typically a directed acyclic graph (DAG) of processing steps -- these systems deliver reproducible execution. 

While the problem of \emph{running} a scientific workflow is, to a large extent, solved, \emph{specifying} a workflow, however, remains manual or semi-automated~\cite{ferreiradasilva2021roadmap}. A~researcher who wants to ``compare mutational patterns in European and African populations across chromosomes 1 through 5'' must translate this intent into a~concrete DAG: locate the correct VCF data files, determine an appropriate parallelism level for the target cluster, and prepare configuration files for the workflow generator script. This translation requires both domain expertise (population codes, genomic region conventions) and infrastructure expertise (available vCPUs, task sizing). In practice, it is performed manually and the translation logic -- the reasoning that connects ``European population'' to specific data files -- must be reconstructed from dataset documentation each time a workflow is specified, rather than encoded in a reusable, auditable form.

This semantic gap between research intent and workflow specification has three consequences. First, it creates a barrier to entry: scientists who lack infrastructure knowledge cannot use workflow systems without assistance. Second, it is error-prone: incorrect vocabulary mappings (e.g., confusing population codes) propagate into results. Third, it undermines reproducibility: when the translation is implicit, repeating an experiment requires reconstructing undocumented decisions.

Existing tools partially address the problem. Visual composers let scientists assemble steps graphically but still require manual infrastructure configuration. Templates pre-encode common pipelines but cannot handle unanticipated research questions. LLMs can generate code from natural language, but direct workflow generation introduces non-determinism~\cite{yildiz2025llms,alam2025prompt} -- the same prompt may yield different DAGs -- undermining reproducibility~\cite{strickland2025schema}.

We present an agentic architecture that automates this translation while preserving reproducibility, based on a three-layer decomposition:
\begin{itemize}
    \item A \emph{semantic layer} where an LLM interprets natural language into a structured intent -- a parameter set capturing the researcher's request (populations, chromosomes, genomic regions) without committing to execution details.
    \item A \emph{deterministic layer} where validated generators transform intents into executable DAGs. 
    \item A \emph{knowledge layer} where domain experts author \emph{Skills} -- markdown documents encoding vocabulary mappings (``European'' $\rightarrow$ EUR), parameter constraints, and optimization strategies. A geneticist writes population mappings; a data engineer documents staging patterns.
\end{itemize}

The architecture is implemented and demonstrated on the 1000 Genomes population genetics workflow~\cite{1000genomes2015}, using the Hyperflow WMS \cite{balis2016hyperflow} on Kubernetes. 

The contributions of this paper are as follows:
\begin{enumerate}
    \item \textbf{Hybrid agentic architecture.} A three-layer decomposition (semantic, deterministic, knowledge) that confines LLM non-determinism to intent extraction while guaranteeing that identical intents produce identical workflows.
    \item \textbf{Skills as domain-expert-authored knowledge.} Skills are text files written in markdown, require no ML expertise, are version-controlled, and can be directly audited by domain experts. 
    \item \textbf{End-to-end agentic pipeline.} A complete path from natural language to Kubernetes execution via four agents (Conductor, Workflow Composer, Deployment Service, Execution Sentinel) with human-in-the-loop validation and deferred workflow generation.
    \item \textbf{Skill-driven execution-time optimization.} Skills encode not only vocabulary for correct translation but also optimization strategies applied at execution time, e.g., selective data extraction that reduced transfer by 92\%.
\end{enumerate}

The paper is organized as follows. \Cref{sec:related-work} discusses related work. \Cref{sec:architecture} presents the architecture -- the three-layer decomposition, pipeline agents, and their interactions. \Cref{sec:skills} details the Skill format and its dual-purpose design. \Cref{sec:evaluation} demonstrates the system on the 1000 Genomes use case. \Cref{sec:conclusion} concludes the paper.


\section{Related Work}
\label{sec:related-work}

Three lines of research converge on the problem we address: LLM-based agents that interpret scientific intent, studies of LLM capabilities for workflows, and knowledge-representation strategies grounding AI reasoning in domain expertise.

\subsection{LLM capabilities for workflow tasks}

Recent empirical studies probe what LLMs can and cannot do with scientific workflows. Yildiz and Peterka~\cite{yildiz2025llms} benchmark four LLMs on workflow configuration, annotation, and cross-platform translation across five HPC systems (Pegasus, Parsl, Swift/T, Balsam, Colmena). Zero-shot performance is poor; few-shot prompting raises configuration accuracy from roughly 30\% to 80\%, but cross-platform translation remains unreliable. Alam and Roy~\cite{alam2025prompt} evaluate GPT-4o, Gemini~2.5 Flash, and DeepSeek-V3 on generating Galaxy and Nextflow bioinformatics workflows. With few-shot prompting, GPT-4o reaches 93\% structural accuracy on Galaxy workflows -- yet no generated workflow executes without manual correction.

Both studies confirm a pattern: LLMs produce structurally plausible workflows but lack the domain vocabulary and platform conventions needed for executable output. Few-shot examples injected at prompt time partially compensate, but the knowledge is ephemeral -- tied to a single prompt rather than maintained as a persistent, auditable artifact. Our Skills address this gap: they encode the same kind of domain knowledge (vocabulary mappings, parameter constraints) as persistent, version-controlled documents that any invocation can consult, rather than examples that must be re-injected into every prompt.

\subsection{Agentic architectures for science}

Agentic AI systems that go beyond single-turn generation have been surveyed from several angles. Ren~\cite{ren2025scientific} propose a mechanism-oriented taxonomy with four components: Planner, Memory, Action Space, Verifier, and catalog agents spanning drug discovery, theorem proving, and materials design. Our Conductor maps to their ``instructional planner'' category; our Skills map to ``external knowledge base'' in their memory taxonomy, though with the distinction that Skills are expert-authored rather than learned or retrieved. Their framework assumes LLM involvement throughout the pipeline -- the fact that our deterministic layer excludes the LLM entirely from workflow generation has no clean mapping in their taxonomy. Gridach~\cite{gridach2025agentic} distinguish fully autonomous systems (closed-loop, minimal human intervention) from human-AI collaborative systems where the scientist retains decision authority. Our architecture falls in the collaborative category: the Conductor enforces human validation gates before provisioning and execution.

Strickland et al.~\cite{strickland2025schema} address the tension between conversational flexibility and execution determinism. Their schema-gated orchestration pattern separates ``conversational authority'' (an LLM interprets user intent) from ``execution authority'' (validated schemas gate every action). Our semantic/deterministic layer split pursues the same goal through a different mechanism: rather than validating LLM output against a schema post hoc, we restrict the LLM to producing a structured intent and delegate workflow generation to deterministic code. 
A further difference: our knowledge layer (Skills) has no counterpart in their framework -- they treat domain knowledge as implicit in schema definitions, while we externalize it as auditable, expert-authored documents.

So~\cite{so2025mcp} examines the Model Context Protocol (MCP) as a standardized interface between LLM agents and scientific tools. MCP addresses tool discovery and invocation -- the ``how'' of agent-tool interaction -- but not the semantic translation from research question to structured intent. MCP servers could complement our architecture as an implementation substrate for the Deployment Service or Execution Sentinel, while our Skills fill a gap MCP does not address: encoding how domain scientists conceptualize their problems.

\subsection{Knowledge representation for agentic systems}

The knowledge-grounding strategies in the literature form a spectrum. At one end, few-shot prompting~\cite{yildiz2025llms,alam2025prompt} injects examples into the LLM context at query time -- effective but ephemeral and not independently auditable. Schema definitions~\cite{strickland2025schema} and MCP tool specifications~\cite{so2025mcp} encode operational constraints but not domain semantics. Retrieval-augmented generation (RAG) retrieves relevant documents from a corpus, but the relevance ranking is itself non-deterministic and the retrieved content is not curated for a specific workflow domain.

Our Skills occupy a distinct point on this spectrum. They are authored by domain scientists in markdown -- a format familiar from documentation and lab notebooks -- and encode vocabulary mappings, parameter constraints, and optimization strategies. No embeddings, training data, or ML expertise are required. The agent consults Skills deterministically (by domain routing, not similarity search), and the same Skill serves both semantic interpretation (``European'' $\rightarrow$ EUR) and operational optimization (data staging patterns). This dual-purpose design has no direct counterpart in the surveyed literature.


\section{Architecture}
\label{sec:architecture}

\begin{figure}[!htb]
\centering
\includegraphics[width=\columnwidth]{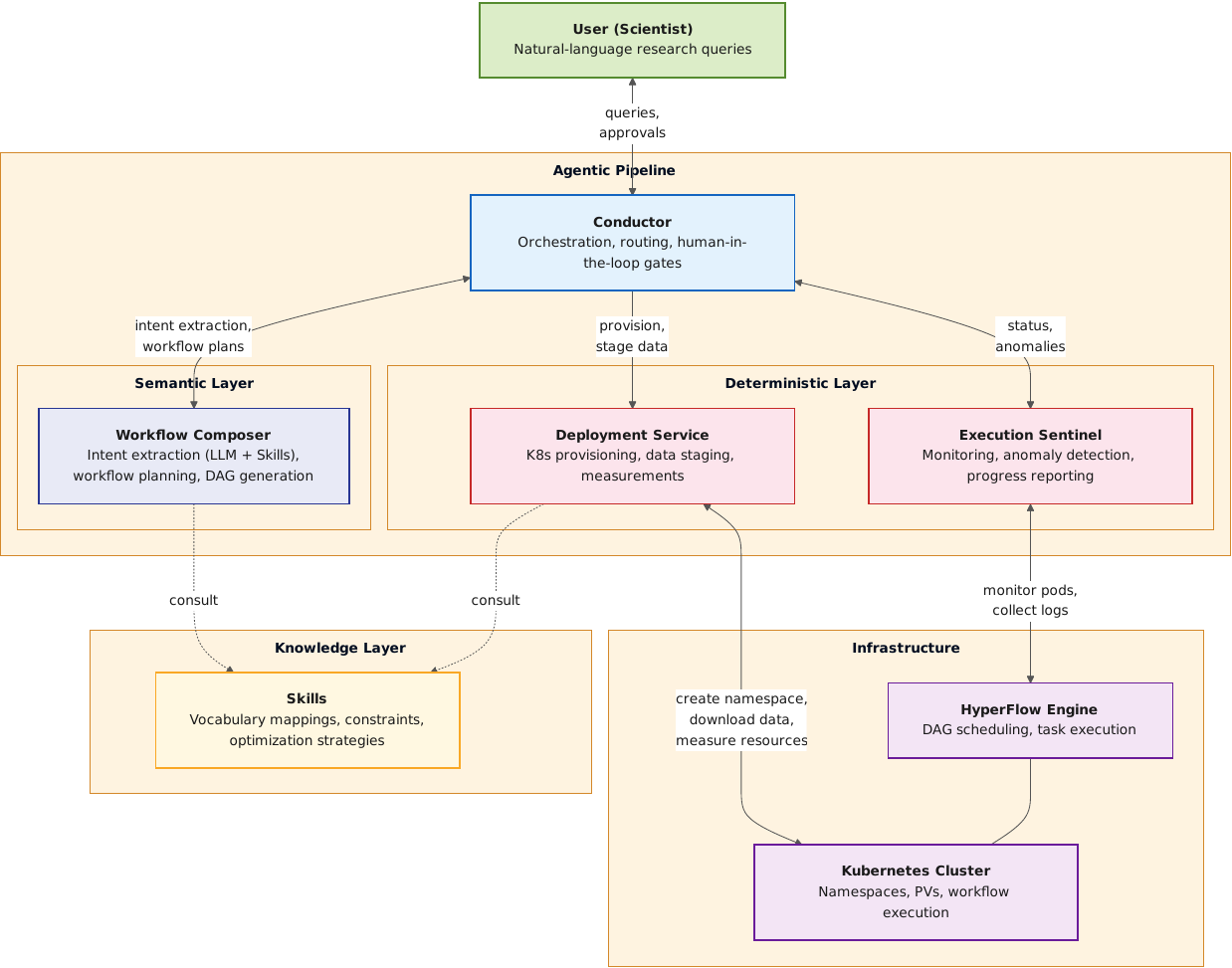}
\caption{Component architecture. The Conductor orchestrates three specialized agents. The Workflow Composer (semantic layer) consults domain Skills (knowledge layer) to produce workflow plans that include data preparation commands. The Deployment Service and Execution Sentinel (deterministic layer) execute these plans on the Kubernetes infrastructure running the HyperFlow engine.}
\label{fig:architecture}
\end{figure}

The system translates natural-language research questions into executable scientific workflows through a pipeline of cooperating agents. \Cref{fig:architecture} shows the component architecture; \cref{fig:pipeline} shows the end-to-end interaction sequence. A~scientist submits a research query and the system returns a running workflow on Kubernetes -- or stops at any point where human judgment is required.

\subsection{Components}
\label{sec:implementation}

The architecture comprises four agents, an external infrastructure layer, and a knowledge layer.

The \emph{Conductor} is the user-facing entry point and orchestrator. It receives natural-language queries, classifies their domain, routes them to the appropriate Workflow Composer, manages multi-turn conversation (clarifications, corrections), and enforces human-in-the-loop validation gates before provisioning and execution. The scientist interacts only with the Conductor and never directly with infrastructure.

The \emph{Workflow Composer} handles intent interpretation and DAG generation. In the planning phase, it consults domain Skills and an LLM to extract a structured intent from the natural-language query -- mapping, for example, ``European population'' to the 1000 Genomes code EUR. It returns a workflow plan with a human-readable description. Later, after infrastructure provisioning, it receives actual measurements (data sizes, available vCPUs) and generates the final \texttt{workflow.json} -- a HyperFlow DAG with resolved parallelism levels and resource allocations.

The key contract between layers is the \texttt{ResearchIntent} -- a schema that the LLM's output must conform to:
\begin{lstlisting}[caption={ResearchIntent schema (simplified).},label={lst:intent},basicstyle=\ttfamily\footnotesize]
ResearchIntent:
  analysis_type: single_population | population_comparison
                 | multi_population | region_analysis
  populations:   list[PopulationCode]   # e.g., [EUR, AFR]
  chromosomes:   list[str] | null
  regions:       list[GenomicRegion] | null
  focus:         all_variants | deleterious | common | rare
\end{lstlisting}

The \emph{Deployment Service} provisions the execution environment. Given an approved plan, it creates a Kubernetes namespace, downloads input data to a cluster persistent volume, and measures actual data sizes and available vCPUs. These measurements feed back into the Workflow Composer for deferred DAG generation -- a design choice that grounds workflow parameters in infrastructure reality rather than estimates.

The \emph{Execution Sentinel} monitors running workflows. It tracks pod status, collects logs, detects anomalies (stalled tasks, repeated failures), and reports progress and completion summaries back to the Conductor. The sequence diagram (\cref{fig:pipeline}) omits this actor for compactness, as monitoring runs asynchronously after workflow submission.

The \emph{knowledge layer} consists of Skills -- markdown documents authored by domain experts -- that the Workflow Composer consults during both intent extraction and data preparation planning. The resulting plan includes concrete staging commands, so the Deployment Service executes it without consulting Skills directly. \Cref{sec:skills} details the format and dual-purpose design of Skills.

These components map onto three architectural layers introduced in \cref{sec:introduction}: the Workflow Composer's LLM-based interpretation constitutes the \emph{semantic layer}, the validated DAG generator and Deployment Service form the \emph{deterministic layer}, and Skills constitute the \emph{knowledge layer}. LLM non-determinism is confined to the semantic layer; once an intent is fixed, the resulting workflow is fully determined. The structured intent, the Skills consulted, and the infrastructure measurements together provide complete composition provenance -- enabling researchers to audit how a query became a specific DAG.

\subsection{Pipeline interaction}
\label{sec:pipeline}

\begin{figure*}[!htb]
\centering
\includegraphics[width=\textwidth]{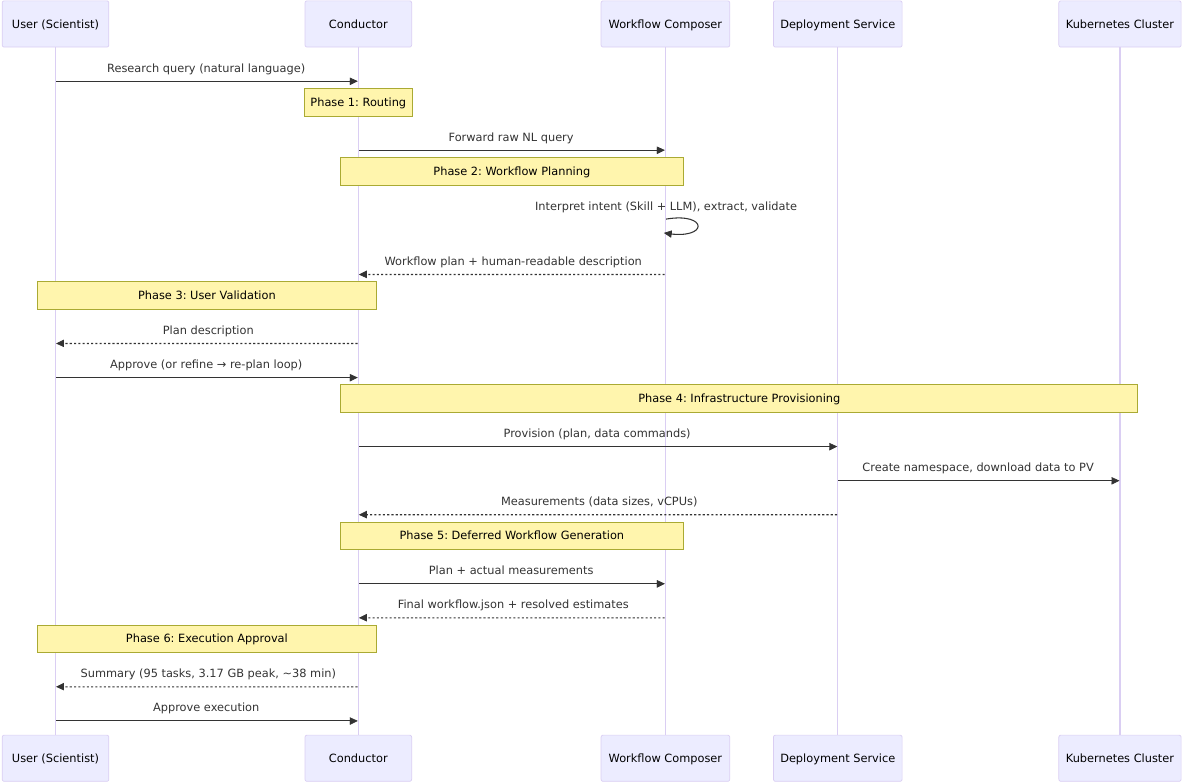}
\caption{Sequence diagram of the agentic pipeline. Five actors -- User, Conductor, Workflow Composer, Deployment Service, and Kubernetes cluster -- interact across six phases. The Execution Sentinel runs asynchronously after workflow submission and is omitted for compactness.}
\label{fig:pipeline}
\end{figure*}

\Cref{fig:pipeline} traces the interaction through six phases:
\begin{enumerate}
\item{\textit{Routing}}. The Conductor receives a natural-language research query and classifies its domain to select the appropriate Workflow Composer and associated Skills.
\item{\textit{Workflow planning}.} The Workflow Composer interprets the query using Skills and an LLM, extracting a structured intent (populations, chromosomes, regions) and producing a workflow plan. Ambiguous queries trigger a clarification loop through the Conductor.
\item{\textit{User validation}.} The Conductor presents the plan for review. The scientist may approve, revise, or reject. Revisions trigger a loop: the Conductor merges corrections with the original context and resubmits to the Composer.
\item{\textit{Infrastructure provisioning}.} Upon approval, the Deployment Service provisions the Kubernetes environment: namespace creation, data download to persistent storage, and measurement of actual data sizes and vCPUs.
\item{\textit{Deferred workflow generation}.} The Deployment Service returns measurements to the Composer, which calibrates parallelism and generates the final DAG. This deferral is deliberate: task parallelism depends on infrastructure state only known after provisioning.
\item{\textit{Execution approval}.} The Conductor presents a summary (task count, estimated peak storage, projected runtime) to the scientist. Upon approval, the workflow is submitted for execution.
\end{enumerate}


\section{Skills}
\label{sec:skills}

The knowledge layer consists of \emph{Skills} -- markdown documents encoding domain expertise for both humans and agents. Each addresses a specific concern: vocabulary mappings, parameter constraints, data source locations, or optimization strategies. Domain experts author Skills in the format they already use for documentation.

\paragraph{Structure.}
In our 1000 Genomes implementation, five Skills cover the knowledge needed:

\begin{itemize}
  \item \textbf{Populations} -- maps natural-language terms (``European,'' ``African ancestry,'' ``Yoruba'') to valid 1000 Genomes codes (EUR, AFR, YRI) at both super-population and sub-population levels, with sample counts.
  \item \textbf{Genomic regions} -- maps gene names and disease contexts (``HLA region,'' ``breast cancer genes'') to GRCh37 coordinates (e.g., chr6:28477797--33448354 for HLA).
  \item \textbf{Research contexts} -- connects high-level research topics (``autoimmune disease,'' ``pharmacogenomics'') to specific regions and analysis types, bridging the gap between scientific intent and workflow parameters.
  \item \textbf{Data sources} -- documents data locations across cloud providers (HTTPS, S3, GCS), extraction patterns (full download vs.\ tabix region extraction), and transfer size estimates that guide staging decisions.
  \item \textbf{Workflow Composer} -- defines available tools, their parameters, and interpretation guidelines that instruct the LLM on how to extract structured intents from natural language.
\end{itemize}

Each document uses definition tables for vocabulary, synonym mapping tables, and decision logic. For example, the populations Skill maps 26 codes to full names and sample counts, with a synonym table resolving ``British'' to GBR and ``Han Chinese'' to CHB.

\paragraph{Dual purpose.} 
Skills serve two roles. First, \emph{correct translation}: for ``compare European and African populations in the HLA region,'' the Composer consults three Skills to resolve EUR/AFR, obtain HLA coordinates, and infer a population-comparison analysis type. Without these mappings, the LLM relies on training data that may be incomplete or hallucinated.

Second, \emph{optimization strategies}: the data sources Skill specifies that tabix extraction of the HLA region transfers ~50~MB vs.\ 943~MB for the full chromosome~6 VCF. The Composer uses this to select the extraction pattern, reducing transfer and storage without manual intervention.


\paragraph{Properties.}
Three properties follow from the design. \emph{Transparency}: Skills are plain markdown that anyone can read and verify, unlike embeddings or model weights. \emph{Version control}: Skills live alongside workflow code in the same repository, so changes are tracked, diffed, and reviewed through standard development workflows. \emph{Domain ownership}: the experts who understand each aspect of the pipeline write and maintain the relevant Skills, without depending on ML engineers to retrain models or rebuild vector stores. Adding support for a new genomic region or population requires editing a markdown table -- a change that takes effect immediately.


\section{Evaluation}
\label{sec:evaluation}

We evaluate on the 1000 Genomes variant-calling workflow~\cite{1000genomes2015}. Its fully enumerable parameter space -- 26 population codes, 5 super-populations, 24 chromosomes, 8 named regions -- makes systematic evaluation tractable.
Intent extraction is evaluated with three models: GPT-5.4, GPT-4.1-mini, and Claude Opus 4.6.
The end-to-end experiments use Gemini 2.0 Flash on a 3-node Kubernetes cluster (48~vCPUs, 165~GB RAM total, Ubuntu 24.04).
We address three questions, each tied to a contribution of the paper.

\paragraph{Query dataset.} We construct 150 natural-language queries with ground-truth \texttt{ResearchIntent} annotations, stratified across five tiers (\cref{tab:query-tiers}). Queries were generated using Claude Opus 4.6 with the complete parameter space and tier-specific rules; the generation script is available for reproducibility. A genomics specialist verified the dataset.

\begin{table}[tb]
\centering 
\caption{Query dataset stratification across five difficulty tiers.}
\label{tab:query-tiers}
\begin{tabular}{@{}clcp{5.5cm}@{}}
\toprule
Tier & Description & Count & Example \\
\midrule
T1 & Explicit (exact codes) & 30 & ``Compare EUR and AFR on chromosome 21'' \\
T2 & Synonym (common names) & 30 & ``Find rare variants in British individuals on chromosome 3'' \\
T3 & Implicit (domain inference) & 30 & ``Profile pharmacogenomic variation across South Asians ethnic groups'' \\
T4 & Underspecified (missing params) & 30 & ``Check TP53 for mutations'' \\
T5 & Adversarial (invalid terms) & 30 & ``Study rare variants in the HBP gene for Mende and Esan populations'' \\
\bottomrule
\end{tabular}
\end{table}


\subsection{Intent extraction accuracy and Skills ablation (C1, C2)}
\label{sec:exp-ablation}

We run intent extraction on all 150 queries in four Skill configurations: (S0)~no Skills loaded (LLM parametric knowledge only), (S1)~vocabulary Skills only (\emph{populations} and \emph{genomic-regions}), (S2)~strategy Skills only (\emph{data-sources} and \emph{research-contexts}), and (S3)~all five Skill documents. Each configuration is evaluated with GPT-5.4, GPT-4.1-mini, and Claude Opus. For each query, the Skill loader is replaced with one that injects only the Skills for the active configuration.

Per-field accuracy (populations, chromosomes, regions) uses exact set matching. Regions are compared as coordinate tuples \texttt{(name, chromosome, start, end)}. The primary metric is full-match accuracy: all fields correct simultaneously. \Cref{tab:ablation-overall} shows results across configurations and models.

\begin{table}[tb]
\centering
\caption{Full-match accuracy (\%) across Skill configurations and models.}
\label{tab:ablation-overall}
\begin{tabular}{@{}lcccc@{}}
\toprule
Model & S0 (none) & S1 (vocab) & S2 (strategy) & S3 (all) \\
\midrule
Claude Opus 4.6 & 44.0 & 80.0 & 57.3 & \textbf{83.3} \\
GPT-5.4         & 39.3 & 78.7 & 48.7 & \textbf{80.0} \\
GPT-4.1-mini    & 27.4 & \textbf{70.7} & 40.0 & 62.0 \\
\bottomrule
\end{tabular}
\end{table}

Vocabulary Skills (S1) drive most of the gain (+36pp for Opus). All models reach 100\% on T1 and T2 with S1, up from 37--80\% -- the population and coordinate tables are decisive.
Strategy Skills alone (S2) give a moderate boost (+9--13pp over S0) but without the coordinate lookup table, region extraction remains unreliable.
The intent extraction performs best with all Skills loaded (with the exception of GPT-4.1-mini).

\paragraph{Per-tier breakdown.} \Cref{tab:ablation-tiers} shows the per-tier full-match accuracy for the S0 (baseline) and S3 (full Skills) configurations.

\begin{table}[tb]
\centering
\caption{Full-match accuracy (\%) by difficulty tier for S0 (no Skills) and S3 (all Skills).}
\label{tab:ablation-tiers}
\begin{tabular}{@{}l cc cc cc@{}}
\toprule
& \multicolumn{2}{c}{GPT-4.1-mini} & \multicolumn{2}{c}{GPT-5.4} & \multicolumn{2}{c}{Claude Opus} \\
\cmidrule(lr){2-3} \cmidrule(lr){4-5} \cmidrule(lr){6-7}
Tier & S0 & S3 & S0 & S3 & S0 & S3 \\
\midrule
T1 (explicit)    & 56.7 & 100.0 & 60.0 & 100.0 & 66.7 & \textbf{100.0} \\
T2 (synonym)     & 36.7 & 100.0 & 70.0 & 100.0 & 80.0 & \textbf{100.0} \\
T3 (implicit)    &  0.0 &  70.0 &  0.0 &  63.3 & 10.0 & \textbf{86.7} \\
T4 (underspecified) & 36.7 &  13.3 & 40.0 &  \textbf{76.7} & 43.3 & 73.3 \\
T5 (adversarial)    &  6.7 &  26.7 & 26.7 &  \textbf{60.0} & 20.0 & 56.7 \\
\midrule
Overall          & 27.4 &  62.0 & 39.3 &  80.0 & 44.0 & \textbf{83.3} \\
\bottomrule
\end{tabular}
\end{table}

With S3, T1 and T2 are solved: all models reach 100\% when queries use explicit codes or common synonyms. The harder tiers separate the models. On T3 (implicit domain inference, e.g., ``breast cancer susceptibility'' $\rightarrow$ BRCA1, BRCA2), Opus reaches 86.7\% while GPT-5.4 and GPT-4.1-mini score 63--70\%. Without any Skills (S0), T3 accuracy is 0--10\% across all models, since parametric knowledge alone cannot produce the exact GRCh37 coordinates the ground truth requires.

\subsection{Deferred Generation Impact (C4)}
\label{sec:exp-deferred}

We compare the advisory plan (Phase~2, estimate-based) against the definitive DAG (Phase~6, measurement-based) across 6 genomic regions from the E2E queries. The results in \cref{tab:deferred-gen} show two effects: parallelism calibration and data transfer savings.

\begin{table}[t]
\centering
\caption{Deferred generation: advisory estimates vs.\ actual measurements. $J$ denotes parallelism (individual jobs per chromosome).}
\label{tab:deferred-gen}
\begin{tabular}{@{}llrrrrr@{}}
\toprule
Region & Chr & Actual rows & Advisory $J$ & Final $J$ & Full VCF & Downloaded \\
\midrule
HLA & 6 & 166,052 & 100 & 51 & 13 GB & 1.57 GB (88\%) \\
BRCA1 & 17 & 2,369 & 100 & 1 & 1.3 GB & 23 MB (98\%) \\
BRCA2 & 13 & 2,502 & 32 & 10 & 1.6 GB & 24 MB (98\%) \\
CFTR & 7 & 4,391 & 66 & 10 & 2.6 GB & 43 MB (98\%) \\
HBB & 11 & 136 & 66 & 1 & 2.1 GB & 1.4 MB (99.9\%) \\
APOE & 19 & 113 & 66 & 1 & 1.0 GB & 1.1 MB (99.9\%) \\
\bottomrule
\end{tabular}
\end{table}

\paragraph{Parallelism calibration.} Without actual row counts, the advisory plan defaults to high parallelism ($J = 32$--$100$). After measuring the extracted VCF files, Phase~6 calibrates $J$ to the actual data volume. For HBB (136 rows) and APOE (113 rows), $J$ drops from 66 to 1 -- generating 66 parallel tasks for 136 rows would waste resources. For HLA (166K rows), $J$ drops from 100 to 51, matching the actual data volume. Without deferred generation, the system would create unnecessary tasks for small regions.

\paragraph{Data transfer savings.} Tabix region extraction downloads only the relevant portion of each chromosome's VCF file. Across the 6 regions, the system downloaded 1.69~GB instead of 21.6~GB -- a 92\% reduction overall. For small gene regions (HBB, APOE), savings exceed 99.9\%: 1.1--1.4~MB instead of 1.0--2.1~GB per chromosome. These savings are specific to region-targeted queries. Whole-chromosome analyses download full VCF files regardless, though parallelism calibration still applies.

\subsection{End-to-End Demonstration (C3)}
\label{sec:exp-e2e}

We execute the complete pipeline\footnote{The query dataset, Skills, and evaluation scripts are available at {\footnotesize \url{https://github.com/hyperflow-wms/1000genome-workflow/tree/main/workflow-composer}}.
} -- from natural-language query to completed workflow execution on Kubernetes -- for 3 queries of increasing complexity (\cref{tab:e2e-results}). Each query targets different genomic regions and populations, testing vocabulary mapping across tiers T2--T4. The three queries are: Q1~=~``Compare HLA and BRCA1 variants in European, African, and East Asian populations''; Q2~=~``Analyze BRCA2 and BRCA1 in British and Finnish populations''; Q3~=~``Compare sickle cell, cystic fibrosis, and Alzheimer's variants across all five super-populations.''

\begin{table}[t]
\centering
\caption{End-to-end results for three queries on a 3-node cluster (48~vCPUs). Times in seconds; percentages show share of total wall-clock time.}
\label{tab:e2e-results}
\begin{tabular}{@{}lrrr@{}}
\toprule
& Q1: HLA+BRCA1 & Q2: BRCA2+BRCA1 & Q3: CFTR+HBB+APOE \\
& 3 populations & 2 populations & 5 populations \\
\midrule
LLM & 14.5\,s (0.2\%) & 13.0\,s (2.2\%) & 11.2\,s (0.7\%) \\
Provisioning & 245.7\,s (2.8\%) & 96.0\,s (16.0\%) & 95.2\,s (6.1\%) \\
Execution & 8464\,s (97.0\%) & 491\,s (81.8\%) & 1457\,s (93.2\%) \\
\midrule
\textbf{Total wall-clock} & \textbf{145\,min} & \textbf{10\,min} & \textbf{26\,min} \\
Total tasks & 118 & 34 & 68 \\
Failed tasks & 0 & 0 & 0 \\
Data downloaded & 1.59\,GB & 47\,MB & 45\,MB \\
Intent accuracy & 5/5 fields & 5/5 fields & 5/5 fields \\
LLM cost & \$0.0007 & \$0.0008 & \$0.0009 \\
\bottomrule
\end{tabular}
\end{table}

Three observations stand out: first, LLM overhead is negligible and constant: 11-14 seconds regardless of query complexity, at a cost below \$0.001 per query (Gemini 2.0 Flash pricing). The semantic layer adds near-zero latency. Second, execution dominates total time (82-97\%), confirming that the Conductor's overhead does not become a bottleneck. The variation in total time -- 10 minutes for small gene regions vs.\ 145 minutes for HLA's 166K variants -- reflects data volume, not system overhead. Third, intent interpretation was correct on all fields for all three queries: populations, chromosomes, regions, analysis type, and focus matched ground truth in every case. Q1's longer runtime reflects HLA's 166K-row dataset requiring 51 parallel jobs, compared to the small gene regions in Q2--Q3.

\paragraph{Comparison with manual specification.} To contextualize the Conductor's value, we describe what the same task requires without it. Manually specifying Q3 
requires five steps: (1)~look up GRCh37 coordinates for HBB, CFTR, and APOE ($\sim$5 min, requires genomics expertise); (2)~write six \texttt{tabix} extraction commands with correct coordinates and download the data ($\sim$10-15 min, requires bioinformatics tooling knowledge); (3)~count variant rows, prepare population sample files, and choose parallelism ($\sim$5 min); (4)~generate the workflow DAG with correct parameters ($\sim$5 min, requires workflow generator script knowledge); (5)~deploy to Kubernetes via Helm and signal the engine ($\sim$10 min, requires DevOps expertise). An expert familiar with all five steps needs 30-50 minutes. The Conductor completes steps 1-5 in 106 seconds (11\,s LLM + 95\,s infrastructure). More significantly, the manual path requires two types of expertise -- domain knowledge (gene coordinates, population codes) and infrastructure knowledge (tabix, HyperFlow, Kubernetes) -- which typically would involve two experts.

\section{Conclusion}
\label{sec:conclusion}

Translating a research question into a running workflow on a cluster requires vocabulary lookups, data source decisions, parallelism calculations, and infrastructure provisioning -- work that scientists currently perform by hand, reconstructing the same reasoning from documentation each time. We presented a~layered agentic architecture that automates this translation while keeping LLM non-determinism out of the generated workflows.

The key design choice -- separating intent extraction (semantic layer) from workflow generation (deterministic layer) -- yields a useful property: identical intents always produce identical DAGs. Skills, authored by domain experts in markdown, provide the vocabulary and optimization knowledge that makes intent extraction accurate. In the ablation study, vocabulary Skills raised full-match accuracy from 44\% to 83\% for Claude Opus and from 39\% to 80\% for GPT-5.4. Deferred generation -- regenerating the DAG after measuring actual data volumes -- reduced data transfer by up to 92\% and eliminated over-parallelization for small genomic regions. The end-to-end pipeline completed three queries of varying complexity on Kubernetes with correct intent extraction on all fields, LLM overhead below 15 seconds per query, and execution cost under \$0.001 per query.

The main limitation is scope: we demonstrated the architecture on one domain (1000 Genomes). The design is meant to be broadly applicable; however, each new domain requires its own set of Skills and its own deterministic generator. The T3 tier results also revealed a~limit: implicit domain reasoning (mapping disease names to gene coordinates) was unreliable even with Skills, suggesting that either richer Skill formats or more capable models are needed.

Future work includes porting the architecture to additional scientific domains, building tooling that helps domain experts author and validate Skills, and feeding execution telemetry back into the planning phase so that the system learns from its own runs.

{\footnotesize \smallskip \noindent
{\bf Acknowledgements.}
This work is co-financed by the European Union NextGenerationEU, Recovery and Resilience Plan under the SPICE project (09I02-03-V01-00012), and supported by the Ministry of Education and Science funds assigned to AGH University. We thank Kinga Zieli\'nska for the discussions on the evaluation dataset.
}

\bibliographystyle{splncs04}
\bibliography{references}

\end{document}